# *An Analytical Study on Behavior of Clusters Using K Means, EM and K[*] Means Algorithm*

Ms. G. Nathiya ,
Lecturer,Department Of Computer Science,
P.S.G.R Krishnammal College For Women,
Coimbatore,Tamilnadu,India
**nathi_ganesh@yahoo.co.in**

Mrs. S. C. Punitha,
Lecturer,Department Of Computer Science,
P.S.G.R Krishnammal College For Women,
Coimbatore,Tamilnadu,India.

Dr. M. Punithavalli
Director of the Computer Science Department,
Sri Ramakrishna college of Arts and Science for Women,
Coimbatore,Tamilnadu,India.

***Abstract*—**Clustering is an unsupervised learning method that constitutes a cornerstone of an intelligent data analysis process. It is used for the exploration of inter-relationships among a collection of patterns, by organizing them into homogeneous clusters. Clustering has been dynamically applied to a variety of tasks in the field of Information Retrieval (IR). Clustering has become one of the most active area of research and the development. Clustering attempts to discover the set of consequential groups where those within each group are more closely related to one another than the others assigned to different groups. The resultant clusters can provide a structure for organizing large bodies of text for efficient browsing and searching. There exists a wide variety of clustering algorithms that has been intensively studied in the clustering problem. Among the algorithms that remain the most common and effectual, the iterative optimization clustering algorithms have been demonstrated reasonable performance for clustering, e.g. the Expectation Maximization (EM) algorithm and its variants, and the well known kmeans algorithm. This paper presents an analysis on how partition method clustering techniques – EM, K –means and K* Means algorithm work on heartspect dataset with below mentioned features – Purity, Entropy, CPU time, Cluster wise analysis, Mean value analysis and inter cluster distance. Thus the paper finally provides the experimental results of datasets for five clusters to strengthen the results that the quality of the behavior in clusters in EM algorithm is far better than kmeans algorithm and k[*]means algorithm.

## I. INTRODUCTION

CLUSTER analysis divides data into meaningful or useful groups (clusters). If meaningful clusters are the goal, then the resulting clusters should capture the "natural" structure of the data. For example, cluster analysis has been used to group related documents for browsing, to find genes and proteins that have similar functionality, and to provide a grouping of spatial locations prone to earthquakes. However, in other cases, cluster analysis is only a useful starting point for other purposes, e.g., data compression or efficiently finding the nearest neighbors of points. Whether for understanding or utility, cluster analysis has long been used in a wide variety of fields: psychology and other social sciences, biology, statistics, pattern recognition, information retrieval, machine learning, and data mining. Generally, clustering algorithms can be categorized into partitioning methods, hierarchical methods, density-based methods, grid-based methods, and model-based methods.

The objective of this paper is to analyze the impact of clusters and its quality by using partitioning method centroid algorithm k-means, k*means and EM (Expectation-Maximization) algorithm. The paper organized as follows. The Section 2 describes about partition methods. The Section 3 describes about K- means algorithm based on centroids. Section 4 explains about k*means algorithm. The section 5 describes about EM algorithm based on probability model with parameters that describe the probability that an instance belongs to a certain cluster. The section 6 describes the experimental results. The section 7 concludes the paper with fewer discussions.

## II. RELATED WORK

The main objective of clustering is to determine the intrinsic grouping in a set of unlabeled data. Consequently, it is the user who must supply this criterion, in such a way that the result of the clustering will suit their needs and the requirements of the corresponding user [11 and 12]. Clustering algorithms may be classified as listed below:

In the case of **Exclusive Clustering** data are grouped in an exclusive way, such that if there exists a certain datum that belongs to a definite cluster then it could not be included in another cluster.

Example: kmeans algorithm

In the **Overlapping Clustering**, the overlapping clustering, uses fuzzy sets to cluster data, such that each of the point may belong to two or more clusters with different degrees of membership [14]. In this case, data will be connected to an appropriate membership value.

Example: Fuzzy C-means algorithm

In the case of a **Hierarchical Clustering** algorithm it is based on the union between the two nearest clusters in the given dataset [15]. The initial condition is realized by setting every datum as a cluster. Subsequently a few iterations it reaches the final clusters wanted.

Example: Agglomerative algorithm

Finally type of clustering is **Probabilistic Clustering** which uses a completely probabilistic approach.

Example: Mixture of Gaussians algorithm

Partitioning methods are divided into two major subcategories, the centroid and the medoids algorithms. The centroid algorithms represent each cluster by using the gravity centre of the instances [4]. The medoid algorithms represent each cluster by means of the instances closest to the gravity centre. The most well known centroid algorithm is the k-means .The k-means method partitions the data set into k subsets such that all points in a given subset are closest to the same centre. In detail, it randomly selects k of the instances to represent the clusters [13]. Based on the selected attributes, all remaining instances are assigned to their closer centre. K-means then computes the new centers by taking the mean of all data points belonging to the same cluster. The operation is iterated until there is no change in the gravity centres. Expectation-Maximization (EM) algorithm assumes an underlying probability model with parameters that describe the probability that an instance belongs to a certain cluster. The strategy in this algorithm is to start with initial guesses for the mixture model parameters. These values are then used to calculate the cluster probabilities for each instance. These probabilities are in turn used to re-estimate the parameters, and the process is repeated.

## III. K-MEANS ALGORITHM

The K-means algorithm assigns each point to the cluster whose center (also called centroid) is nearest. The center is the average of all the points in the cluster that is, its coordinates are the arithmetic mean for each dimension separately over all the points in the cluster [2]. K-Means can be thought of as an algorithm relying on hard assignment of information to a given set of partitions. At every pass of the algorithm, each data value is assigned to the nearest partition based upon some similarity parameter such as Euclidean distance of intensity [5]. The partitions are then recalculated based on these hard assignments. With each successive pass, a data value can switch partitions, thus altering the values of the partitions at every pass. K-Means algorithms typically converge to a solution very quickly as opposed to other clustering algorithms. The algorithm steps are:

- Choose the number of clusters, *k*.
- Randomly generate *k* clusters and determine the cluster centers, or directly generate *k* random points as cluster centers.
- Assign each point to the nearest cluster center.
- Recompute the new cluster centers.
- Repeat the two previous steps until some convergence criterion is met (usually that the assignment hasn't changed).

This type of algorithm needs the input information of exactly the number of clusters that are to be as distinct as possible. Thus such type of research queries can be addressed by the k- means clustering algorithm. In general, the k-means method will produce exactly k different clusters which have greatest possible distinction [18]. The best number of clusters k leading to the greatest separation among the clusters is not the priori and must be computed from the data.

The result of a k-means clustering analysis, the means for each cluster on each dimension is examined to assess how distinct our k clusters are. Very different means are obtained for most, if not all dimensions, used in this analysis [19]. The magnitude of the values from the analysis of this variance performed on each dimension shows of how well the particular dimension discriminates between clusters. The fundamental process of the algorithm is relatively simple: A fixed number of k clusters are taken into consideration, and then assign observations to those clusters so that the means across clusters are as different from each other as possible.

## IV. K∗ MEANS ALGORITHM

This section of the paper presents a STep-wise Automatic Rival-penalized (STAR) k-means algorithm which is a generalized version of traditional k-means clustering algorithm [2]. The initial step in the k*means clustering algorithm is to let each cluster acquires at least one seed point. This first step can be considered as a pre-processing procedure. Then, the next step is to fine-tune the units adaptively by a learning rule that automatically penalizes the winning chance of all rival seed points in the consequent competitions while tuning the winning one to adapt to an input. The detailed k*means can be given out as follows,

- The initial step is generally implemented using Frequency Sensitive Competitive Learning such that they can achieve the goal as long as the number of seed points is not less than the exact number k* of clusters. Therefore the number of clusters is k ≥ k*, and randomly initialize the k seed points.
- Randomly pick up a data point $x_t$, from the input data set, and for j = 1, 2, . . . k let,

$$U_j = \{1\} \text{ if } j = w = \arg\min_r \lambda_r \|x_t - m_r\| \text{ , 0 otherwise.}$$

Where $\lambda_r = n_j \Big/ \sum_{r=1}^{k} n_r$ and $n_r$ represents the cumulative number of occurrences of $u_r = 1$.

- The next step is to update the winning seed point mw by the following equation,

$$m_w^{new} = m_w^{old} + \eta \; (x_t - m_w^{old})$$

where $\eta$ is the smallest positive learning rate. The input covariance is not included because they merely aims to allocate the seed points into some desired regions, rather than making a precise value estimate of them.

- The above mentioned two steps are repeated until the k series of uj, j = 1, 2, . . . k, remain unchanged for all $x_t$s.
- Initialize $\alpha_j = 1/k$, for j = 1, 2, . . . ,k, and let $\Sigma_j$ be the covariance matrix of those data points with $u_j = 1$.
- Given a data point $x_t$, calculate $I(j|x_t)$s by using the following equation,

  $I(j|x) = \{1$, if $j = w = \arg\min_r \rho_r$, 0 otherwise.
- Update the winning seed point $m_w$ by the following equation,

$$m_w^{new} = m_w^{old} + \eta \sum_w^{-1} (x_t - m_w^{old})$$

  Further the parameters like $\alpha_j$s and $\Sigma_w$ should be updated.
- The above mentioned two steps are repeated until the k series of $I(j|x_t)$, with j = 1, 2, . . . k, remain unchanged for all $x_t$s.

The k*means clustering algorithm overcomes some of the limitations of conventional k means clustering algorithm. The k*means clustering algorithm eliminates the problem of dead-unit that was available with conventional k means clustering algorithm.

## V. EM ALGORITHM

The general purpose of EM techniques is to detect clusters in observations (or variables) and to assign those observations to the clusters. A typical example application for this type of analysis is a marketing research study in which a number of consumer behavior related variables are measured for a large sample of respondents. The purpose of the study is to detect "market segments," i.e., groups of respondents that are somehow more similar to each other (to all other members of the same cluster) when compared to respondents that "belong to" other clusters. In addition to identifying such clusters, it is usually equally of interest to determine how the clusters are different, i.e., determine the specific variables or dimensions that vary and how they vary in regard to members in different clusters.

The fundamental approach and logic of this clustering method is to compute a single continuous large variable in a huge sample of observations [16 and 17]. Further, considering that the sample of given dataset consists of two clusters of observations with different means within each sample, and then the distribution of values for the continuous large variable follow a normal distribution.

The EM (expectation maximization) algorithm extends this basic approach to clustering in two important ways: Instead of assigning cases or observations to clusters to maximize the differences in means for continuous variables, the EM clustering algorithm computes probabilities of cluster memberships based on one or more probability distributions [3]. The goal of the clustering algorithm then is to maximize the overall probability or likelihood of the data, given the (final) clusters. Unlike the classic implementation of k-means clustering, the general EM algorithm can be applied to both continuous and categorical variables (note that the classic k-means algorithm can also be modified to accommodate categorical variables) [1].

The EM algorithm is very similar in setup to the K-Means algorithm. Similarly, the first step is to choose the input partitions. The EM cycle begins with an Expectation step, which is defined by the following equation:

$$E[z_{ij}] = \frac{p(x = x_i \mid \mu = \mu_j)}{\sum_{n=1}^{k} p(x = x_i \mid \mu = \mu_n)}$$

$$= \frac{e^{-\frac{1}{2\sigma^2}(x_i - \mu_j)^2}}{\sum_{n=1}^{k} e^{-\frac{1}{2\sigma^2}(x_i - \mu_n)^2}}$$

This equation states that the expectations or weight for pixel z with respect to partition j equals the probability that x is pixel $x_i$ given that μ is partition $\mu_i$ divided by the sum over all partitions k of the same previously described probability. This leads to the lower expression for the weights. The sigma squared seen in the second expression represents the covariance of the pixel data. Once the E step has been performed and every pixel has a weight or expectation for each partition, the M step or maximization step begins. This step is defined by the following equation:

$$\mu_j \leftarrow \frac{1}{m} \sum_{i=1}^{m} E[z_{ij}] \; x_i$$

This equation states that the partition value j is changed to the weighted average of the pixel values where the weights are the weights from the E step for this particular partition. This EM cycle is repeated for each new set of partitions until, as in the K-Means algorithm, the partition values no longer change by a significant amount.

## VI. EXPERIMENTAL RESULTS

In this section the paper discusses the behavior of the clusters by taking SPECTF dataset into consideration for the clustering of five cluster groups which is done by taking the results of both the k means, k*means and the EM algorithm for the need of comparison to bring out the experimental results. We calculate the percentage of dominant class members after applying the purity measure and thus entropy is calculated. The paper verifies all the most common criteria for the comparison of the datasets that are clustered with the measurement of the CPU time for EM, k*means and k means algorithm. We compute the mean values for the dataset for clustering using both the EM, k*means and k means algorithm. An overall analysis has been carried out to study the cluster quality based on all these main criteria for the values of both the EM, k*means and K-means algorithm. The algorithms have been executed using MATLAB version 7.0.The clusters formed by EM and K-means algorithm has been evaluated.

### A. Heart Spect Dataset

SPECTF is a good data set for testing ML algorithms; it has 267 instances that are described by 45 attributes. The dataset describes diagnosing of cardiac Single Proton Emission Computed Tomography (SPECT) images. Each of the patients is classified into two categories: normal and abnormal. The database of 267 SPECT image sets (patients) was processed to extract features that summarize the original SPECT images. All the inputs in the data set where normalized. It can be scaled in the range of 0 to 1.

TABLE-1

PURITY MEASURE (NORMAL)

| NUMBER OF 1s | | |
|---|---|---|
| K means | K*Means | EM |
| 3403 | 3390 | 3373 |

TABLE-2

CPU TIME

| CPU Time | | | |
|---|---|---|---|
| K-means | K*means | EM | K-means + K*means+ EM |
| 2.7380e+003 | 1.7385e+003 | .7388e+003 | 3.2424e+003 |

TABLE-3

MEAN WISE COMPARISON

| K-means | K*means | EM |
|---|---|---|
| 22.7297 | 21.2556 | 19.5883 |
| 44.7815 | 36.4515 | 29.6015 |
| 58.6818 | 57.9650 | 57.1785 |
| 66.8687 | 65.6565 | 64.7206 |
| 74.8776 | 72.9605 | 71.1364 |

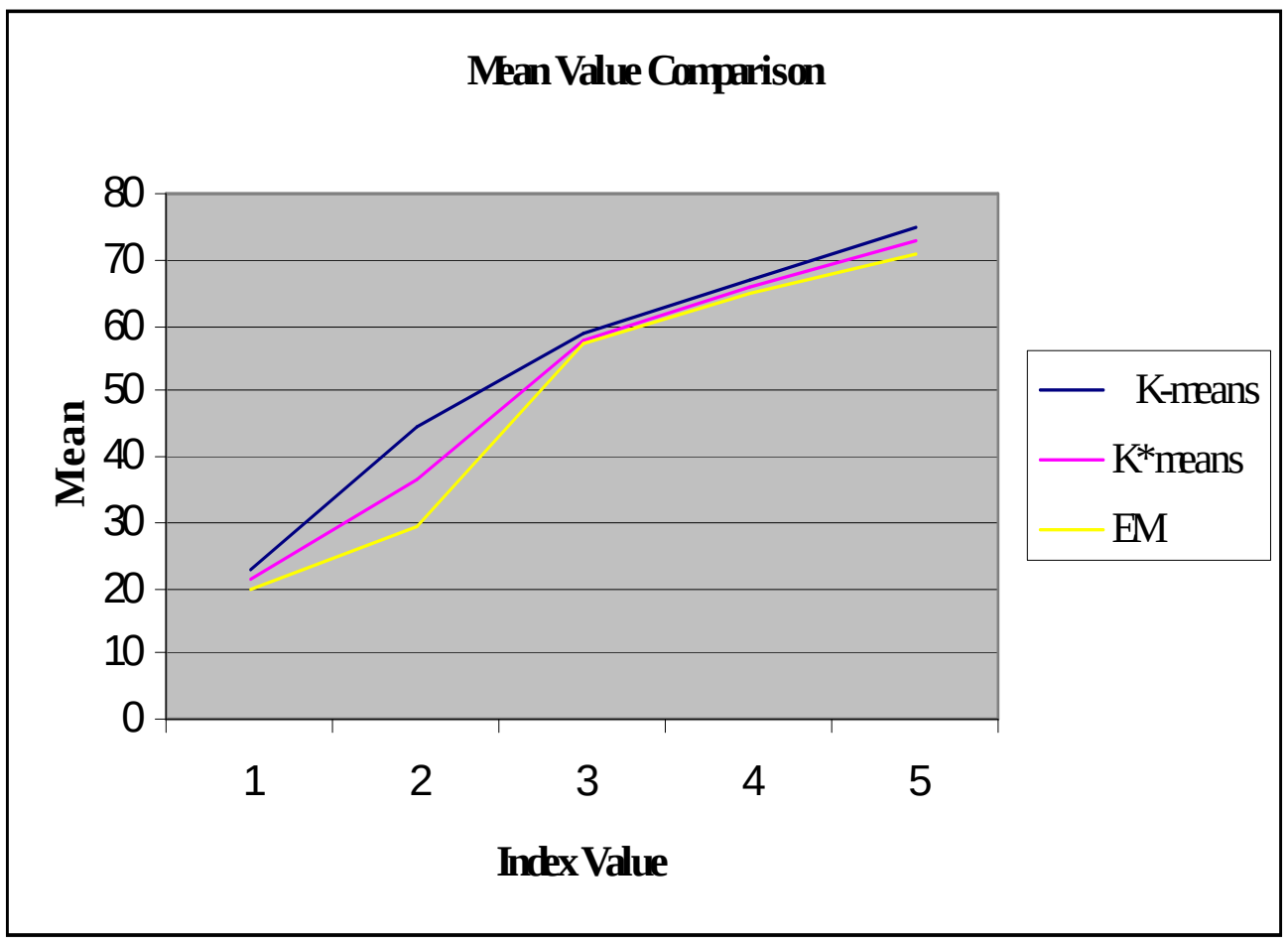

Fig. 1 Results on Comparing the Mean values of EM and K-means algorithm

Purity measures the percentage of the dominant class members in a given cluster (larger is better), while entropy looks at the distribution of documents from each reference class within clusters (smaller is better). These are written as

$$\text{Purity}=\sum_j \frac{n_j}{n} \arg\max_i p(i,j)$$

$$N_{entro} = -\frac{1}{\log K}\sum_{k=1}^{K}\frac{N_k}{N}\log\left(\frac{N_k}{N}\right)$$

The percentage of dominant class members with 1 after applying the purity measures as follows in (Table-1). The entropy is calculated for each cluster type and compared. This section discusses the time consumption of CPU which is one of the most important criterions to be taken into consideration, where the Table-2 summarizes CPU time of EM, K*means, K-means algorithm and the CPU time of combined EM and K-means.

Table-3 shows the mean values of derived from K-means, K*means and EM algorithm for five types of clusters. The graph in the figure 1 shows the experimental results showing the behavior of the clusters in k*means, k means and the EM methodology. Thus, there is clear progress in the EM methodology when compared to the k means algorithm and k*means clustering algorithms.

The cluster limit assigned for dataset is five and the plotting of cluster within these five clusters is indicated with its index and position values. Figure 2 shows the results on the dataset clustering within these five clusters

The graphs in the figure 2 (a, b, c, d,and e) illustraes the behavior of the SPECTF dataset when clustered after first, second, third, fourth and fifth respectively. The graphs are being plotted with the index against their position values of each dataset for five clustering groups in both the algorithms. This clearly shows that in there is a remarkable development in the results brought out by the EM algorithm when compared to the Kmeans algorithm and K*means in each time when the dataset is being clustered.

## VII. CONCLUSION

This paper discusses the partition methods in brief where the k-means and k*means methodology is described in detail. Then demonstrates the cluster quality with purity measure for EM, K*means and K means algorithms Cluster wise analysis for EM, K*means and K- means is analyzed with five clusters. Mean value and CPU time performance is analyzed using K-means, K*means and EM. Finally the experimental results are determined which shows that there is tremendous improvement in the quality of the behavior of clusters in the EM algorithm when compared to the k*means and k means algorithm.

There needs a great deal of future work to be done in the field of clustering technique which is active area of research which have been have been applied to a wide variety of research problems. For example, in the field of medicine for the clustering diseases, to cures for diseases, or to classify the symptoms of diseases that can lead to very useful taxonomies in the field of psychiatry, the correct diagnosis of clusters of symptoms such as paranoia, schizophrenia, etc. is essential for successful therapy. In archeology, researchers have attempted to establish taxonomies of stone tools, funeral objects, etc. by applying cluster analytic techniques. Thus this area has to make many improvements in this field of research. In general, whenever one requires classifying a very large amount of information into manageable meaningful piles, cluster analysis is a great utility.

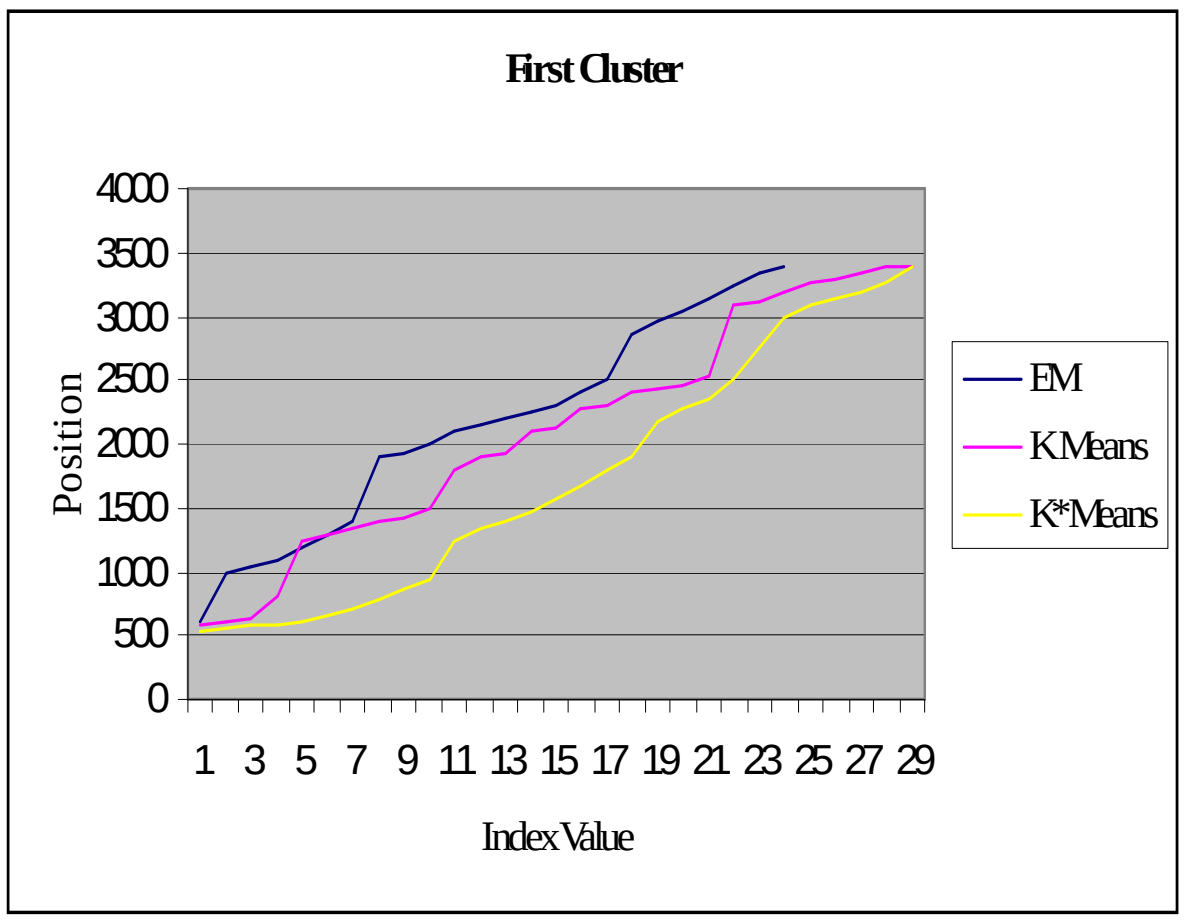

(a)

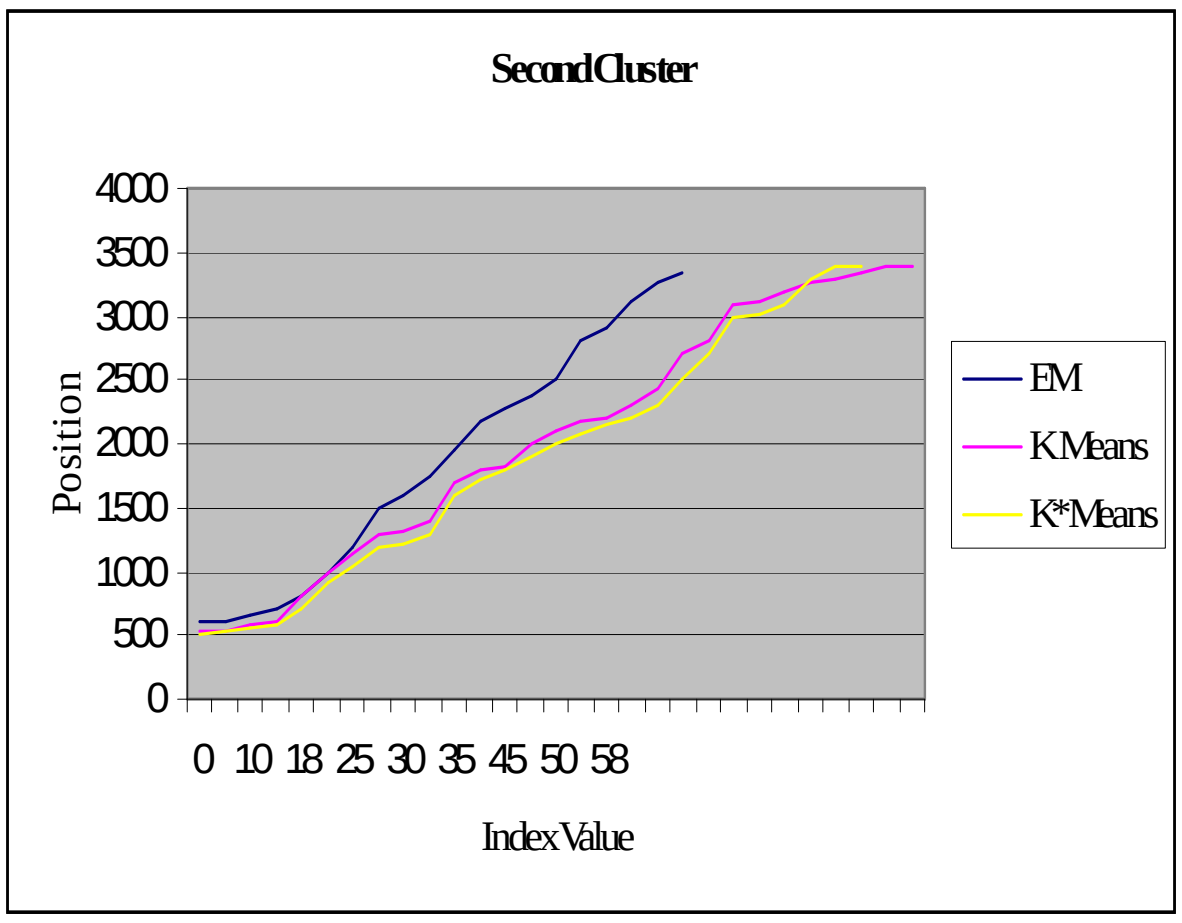

(b)

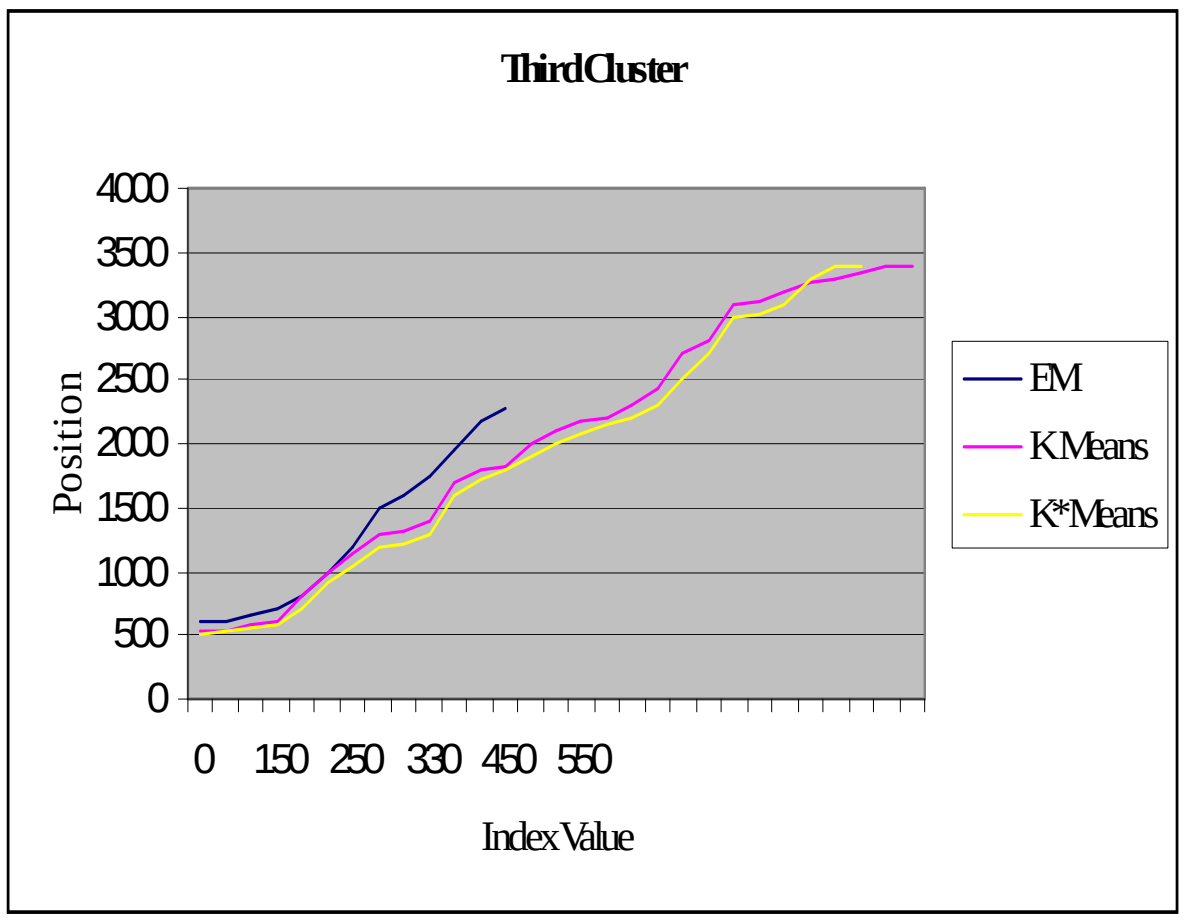

(c)

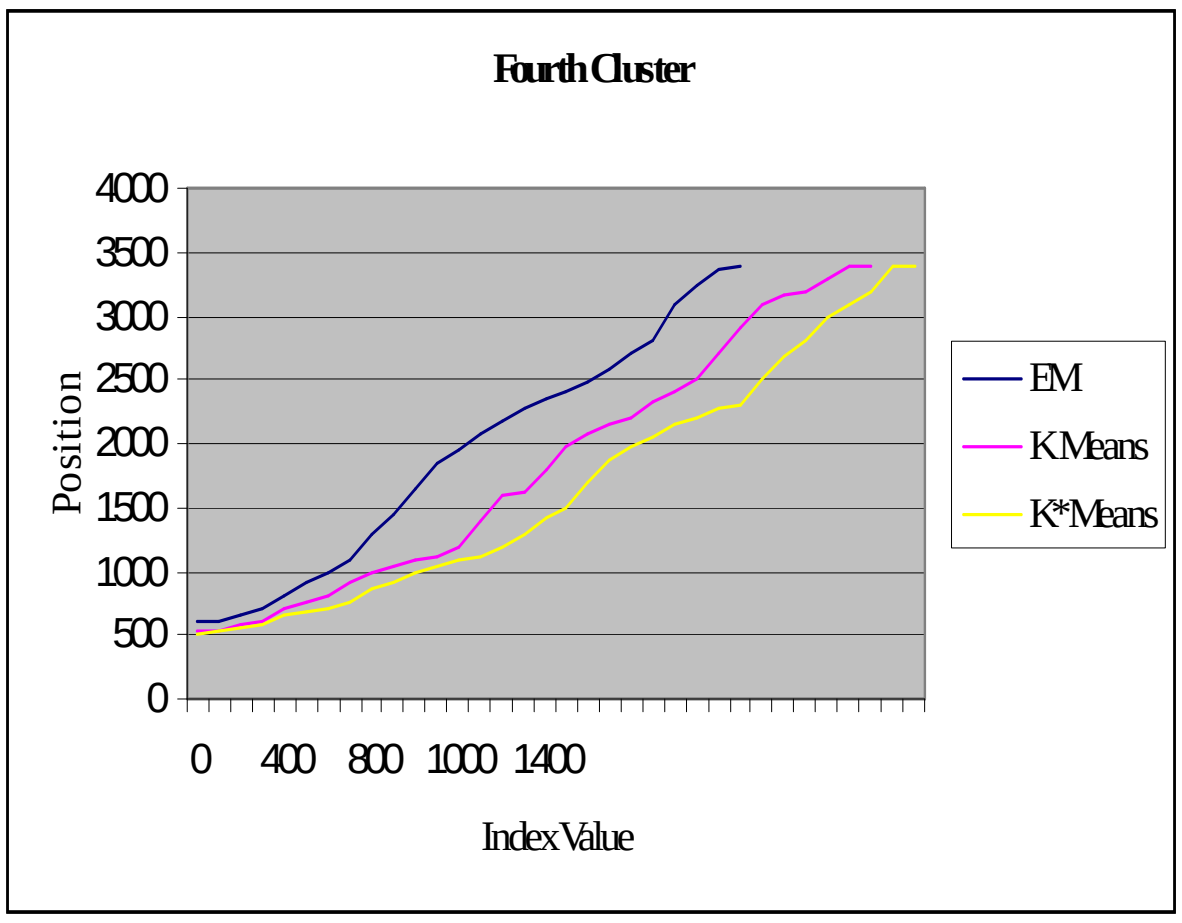

(d)

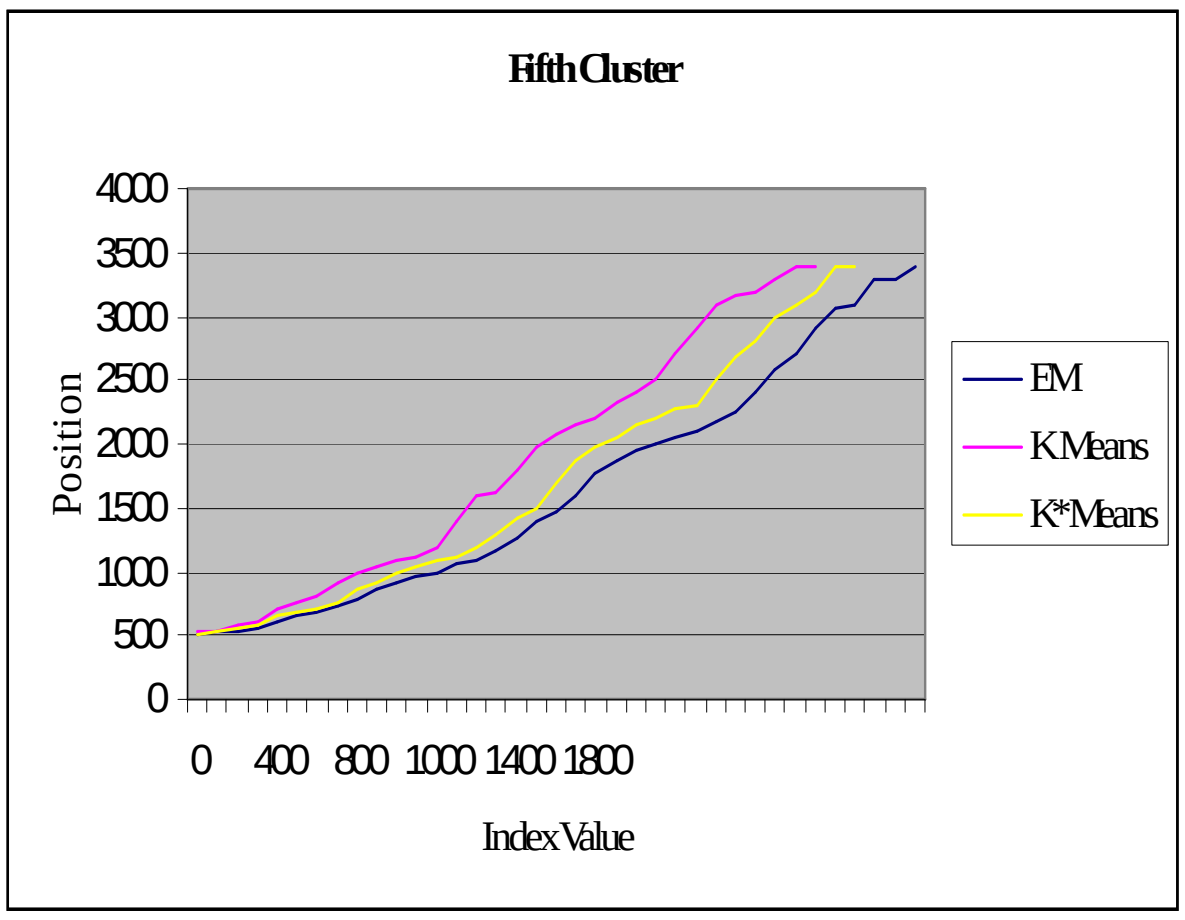

(e)

Fig. 2: Results on SPECTF dataset: Clustering of (a) SPECTF dataset into first cluster (b) SPECTF dataset into second cluster (c) SPECTF dataset into Third cluster (d) SPECTF dataset into Fourth cluster (e) SPECTF dataset into fifth cluster

**Author’s Profile**

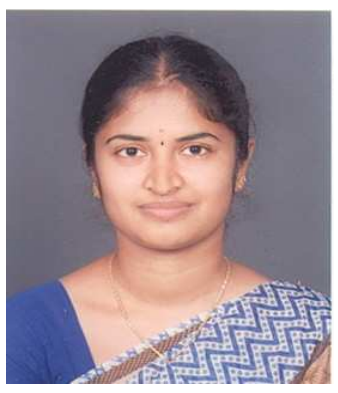

**G. Nathiya,** M.Sc., is a lecturer in the Department of Computer Science and Information Technology, P.S.G.R Krishnammal College for Women, Coimbatore. She completed her BCA in P.S.G.R Krishnammal College for Women, Coimbatore. She completed her M.Sc in Sri Ramakrishna college of arts and science for women, Coimbatore. She is pursuing her M.Phil., in Bharathiar University Coimbatore. Her current research interest includes Data Mining and Neural Networks.

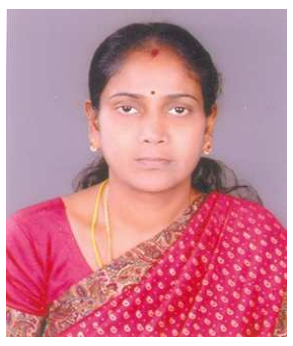

**S. C. Punitha,** HOD, Department of Computer Science and Information Technology, P. S. G. R Krishnammal College for Women, Coimbatore. She completed her M.Sc., MFT., M.Phil., from Bharathiar University. She is pursuing her Ph.D., in Karunya University, Coimbatore in the field of Computer Science. She has an academic experience of 12 years. Her area of research interest includes Data mining and Artificial Intelligence.

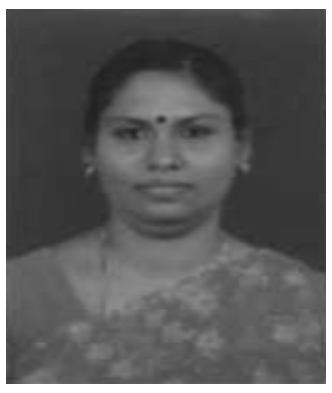

**Dr. M. Punithavalli** received the Ph.D degree in Computer Science from Alagappa University, Karaikudi in May 2007. She is currently serving as the Director of the Computer Science Department, Sri Ramakrishna college of Arts and Science for Women, Coimbatore. Her research interest lies in the area of Data mining, Genetic Algorithms and Image Processing. She has published more than 10 Technical papers in International, National Journals and conferences. She is Board of studies member various universities and colleges. She is also reviewer in International Journals. She has given many guest lecturers and acted as chairperson in conference. Currently 10 students are doing Ph.D under her supervision